\documentclass[runningheads]{llncs}

\usepackage{graphicx}
\usepackage[hidelinks]{hyperref}

\usepackage{amsmath}

\usepackage{booktabs}
\newcommand{\ra}[1]{\renewcommand{\arraystretch}{#1}}

\begin{document}

\title{The Role of Subgroup Separability in Group-Fair Medical Image Classification}
\titlerunning{Subgroup Separability in Medical Image Classification}  

\author{Charles Jones \and M\'elanie Roschewitz \and
Ben Glocker
}

\authorrunning{C. Jones et al.}

\institute{Department of Computing, Imperial College London, UK 
\email{\{charles.jones17,mb121,b.glocker\}@imperial.ac.uk}\\
}
\maketitle              

\begin{abstract}
We investigate performance disparities in deep classifiers. We find that the ability of classifiers to separate individuals into subgroups varies substantially across medical imaging modalities and protected characteristics; crucially, we show that this property is predictive of algorithmic bias. Through theoretical analysis and extensive empirical evaluation\footnote{Code is available at \url{https://github.com/biomedia-mira/subgroup-separability}}, we find a relationship between subgroup separability, subgroup disparities, and performance degradation when models are trained on data with systematic bias such as underdiagnosis. Our findings shed new light on the question of how models become biased, providing important insights for the development of fair medical imaging AI.

\end{abstract}

\section{Introduction}

Medical image computing has seen great progress with the development of deep image classifiers, which can be trained to perform diagnostic tasks to the level of skilled professionals \cite{Rajpurkar2017}. Recently, it was shown that these models might rely on sensitive information when making their predictions \cite{gichoyaAIRecognitionPatient2022,glockerAlgorithmicEncodingProtected2023} and that they exhibit performance disparities across protected population subgroups \cite{seyyed-kalantariUnderdiagnosisBiasArtificial2021}. Although many methods exist for mitigating bias in image classifiers, they often fail unexpectedly and may even be harmful in some situations \cite{zietlowLevelingComputerVision2022}. Today, no bias mitigation methods consistently outperform the baseline approach of empirical risk minimisation (ERM) \cite{vapnikOverviewStatisticalLearning1999,zongMEDFAIRBenchmarkingFairness2023}, and none are suitable for real-world deployment. If we wish to deploy appropriate and fair automated systems, we must first understand the underlying mechanisms causing ERM models to become biased.  

An often overlooked aspect of this problem is \emph{subgroup separability}: the ease with which individuals can be identified as subgroup members. Some medical images encode sensitive information that models may leverage to classify individuals into subgroups \cite{gichoyaAIRecognitionPatient2022}. However, this property is unlikely to hold for all modalities and protected characteristics. A more realistic premise is that subgroup separability varies across characteristics and modalities. We may expect groups with intrinsic physiological differences to be highly separable for deep image classifiers (e.g. biological sex from chest X-ray can be predicted with $> 0.98 \text{ AUC}$). In contrast, groups with more subtle differences (e.g. due to `social constructs') may be harder for a model to classify. This is especially relevant in medical imaging, where attributes such as age, biological sex, self-reported race, socioeconomic status, and geographic location are often considered sensitive for various clinical, ethical, and societal reasons.

We highlight how the separability of protected groups interacts in non-trivial ways with the training of deep neural networks. We show that the ability of models to detect which group an individual belongs to varies across modalities and groups in medical imaging and that this property has profound consequences for the performance and fairness of deep classifiers. To the best of our knowledge, ours is the first work which analyses group-fair image classification through the lens of subgroup separability. Our contributions are threefold:

\begin{itemize}
    \item We demonstrate empirically that subgroup separability varies across real-world modalities and protected characteristics.
    \item We show theoretically that such differences in subgroup separability affect model bias in learned classifiers and that group fairness metrics may be inappropriate for datasets with low subgroup separability.
    \item We corroborate our analysis with extensive testing on real-world medical datasets, finding that performance degradation and subgroup disparities are functions of subgroup separability when data is biased.
\end{itemize}

\section{Related Work} 

Group-fair image analysis seeks to mitigate performance disparities caused by models exploiting sensitive information. In medical imaging, Seyyed-Kalantari et al. \cite{seyyed-kalantariUnderdiagnosisBiasArtificial2021} highlighted that classification models trained through ERM underdiagnose historically underserved population subgroups. Follow-up work has additionally shown that these models may use sensitive information to bias their predictions \cite{glockerAlgorithmicEncodingProtected2023,gichoyaAIRecognitionPatient2022}. Unfortunately, standard bias mitigation methods from computer vision, such as adversarial training \cite{kimLearningNotLearn2019,alviTurningBlindEye2018} and domain-independent training \cite{wangFairnessVisualRecognition2020}, are unlikely to be suitable solutions. Indeed, recent benchmarking on the MEDFAIR suite \cite{zongMEDFAIRBenchmarkingFairness2023} found that no method consistently outperforms ERM. On natural images, Zietlow et al. \cite{zietlowLevelingComputerVision2022} showed that bias mitigation methods worsen performance for all groups compared to ERM, giving a stark warning that blindly applying methods and metrics leads to a dangerous `levelling down' effect \cite{mittelstadtUnfairnessFairMachine2023}.  

One step towards overcoming these challenges and developing fair and performant methods is understanding the circumstances under which deep classifiers learn to exploit sensitive information inappropriately. Today, our understanding of this topic is limited. Closely related to our work is Oakden-Rayner et al., who consider how `hidden stratification' may affect learned classifiers \cite{oakden-raynerHiddenStratificationCauses2020}; similarly, Jabbour et al. use preprocessing filters to inject spurious correlations into chest X-ray data, finding that ERM-trained models are more biased when the correlations are easier to learn \cite{jabbourDeepLearningApplied2020}. Outside of fairness, our work may have broader impact in the fields of distribution shift and shortcut learning \cite{wilesFineGrainedAnalysisDistribution2022,geirhosShortcutLearningDeep2020}, where many examples exist of models learning to exploit inappropriate spurious correlations \cite{degraveAIRadiographicCOVID192021,brownDetectingPreventingShortcut,nautaUncoveringCorrectingShortcut2021}, yet tools for detecting and mitigating the problem remain immature.

\section{The Role of Subgroup Separability} \label{sec:analysis}

Consider a binary disease classification problem where, for each image $x \in X$, we wish to predict a class label $y \in Y: \{y^{+},y^{-}\}$. We denote $P: [Y | X] \to [0,1]$ the underlying mapping between images and class labels. Suppose we have access to a (biased) training dataset, where $P_{tr}$ is the conditional distribution between training images and training labels; we say that such a dataset is biased if $P_{tr} \ne P$. We focus on group fairness, where each individual belongs to a subgroup $a \in A$ and aim to learn a fair model that maximises performance for all groups when deployed on an unbiased test dataset drawn from $P$. We assume that the groups are consistent across both datasets. The bias we consider in this work is underdiagnosis, a form of label noise \cite{castroCausalityMattersMedical2020} where some truly positive individuals $x^+$ are mislabeled as negative. We are particularly concerned with cases where underdiagnosis manifests in specific subgroups due to historic disparities in healthcare provision or discriminatory diagnosis policy. Formally, group $A = a^*$ is said to be underdiagnosed if it satisfies Eq. (\ref{eq:underdiagnosis}):

\begin{equation} \label{eq:underdiagnosis}
    P_{tr}(y \vert x^+, a^*) \le P(y \vert x^+, a^*) \ \text{and} \ \forall a \ne a^*, P_{tr}(y \vert x^+, a) = P(y \vert x^+, a)
\end{equation}

 We may now use the law of total probability to express the overall mapping from image to label in terms of the subgroup-wise mappings in Eq. (\ref{eq:law_total_prob}). Together with Eq. (\ref{eq:underdiagnosis}), this implies Eq. (\ref{eq:underdiagnosis_total}) -- the probability of a truly positive individual being assigned a positive label is lower in the biased training dataset than for the unbiased test set. 
 
\begin{equation} \label{eq:law_total_prob}
     P_{tr}(y \vert x) = \sum_{a \in A}P_{tr}(y \vert x, a)P_{tr}(a \vert x)
\end{equation}

\begin{equation}\label{eq:underdiagnosis_total}
P_{tr}(y \vert x^+) \le P(y \vert x^+)
\end{equation}

At training time, supervised learning with empirical risk minimisation aims to obtain a model $\hat{p}$, mapping images to predicted labels $\hat{y} = \text{argmax}_{y\in Y} \hat{p}(y \vert x)$ such that $\hat{p}(y \vert x) \approx P_{tr}(y \vert x), \forall (x, y)$. Since this model approximates the biased training distribution, we may expect underdiagnosis from the training data to be reflected by the learned model when evaluated on the unbiased test set. However, \emph{the distribution of errors from the learned model depends on subgroup separability}. Revisiting Eq. (\ref{eq:law_total_prob}), notice that the prediction for any individual is a linear combination of the mappings for each subgroup, weighted by the probability the individual belongs to each group.  When subgroup separability is high due to the presence of sensitive information, the model will learn a different mapping for each subgroup, shown in Eq. (\ref{eq:sep}) and Eq. (\ref{eq:recoveredmapping}). This model underdiagnoses group $A=a^*$ whilst recovering the unbiased mapping for other groups.

\begin{equation}\label{eq:sep}
    \hat{p}(y \vert x^+, a^*) \approx P_{tr}(y \vert x^+, a^*) \le P(y \vert x^+, a^*)
\end{equation}

\begin{equation} \label{eq:recoveredmapping}
    \text{and} \ \forall a \ne a^*, \ \hat{p}(y \vert x^+, a) \approx P_{tr}(y \vert x^+, a) = P(y \vert x^+, a)
\end{equation}

Eq. (\ref{eq:sep}) and Eq. (\ref{eq:recoveredmapping}) show that, at test-time, our model will demonstrate worse performance for the underdiagnosed subgroup than the other subgroups. Indeed, consider True Positive Rate (TPR) as a performance metric. The group-wise TPR of an unbiased model, $\text{TPR}_{a}^{(u)}$, is expressed in Eq. (\ref{eq:tpr_unbiased}). 

\begin{equation} \label{eq:tpr_unbiased}
    \text{TPR}_{a}^{(u)} = \frac{|\hat{p}(y|x^{+}, a) > 0.5|}{N_{+, a}} \approx \frac{|P(y|x^{+}, a) > 0.5|}{N_{+, a}}
\end{equation}

Here, $N_{+, a}$ denotes the number of positive samples belonging to group $a$ in the test set. Remember, in practice, we must train our model on the biased training distribution $P_{tr}$. We thus derive test-time TPR for such a model, $\text{TPR}_{a}^{(b)}$, from Eq. (\ref{eq:sep}) and Eq. (\ref{eq:recoveredmapping}), giving Eq. (\ref{eq:tpr_sep}) and Eq. (\ref{eq:tpr_recovered}).    

\begin{equation} \label{eq:tpr_sep}
    \text{TPR}_{a^*}^{(b)} \approx \frac{|P_{tr}(y|x^{+}, a^*) > 0.5|}{N_{+, a^*}} \le \frac{|P(y|x^{+}, a^*) > 0.5|}{N_{+, a^*}} \approx \text{TPR}_{a^*}^{(u)}
\end{equation}

\begin{equation} \label{eq:tpr_recovered}
    \text{and} \ \forall a \ne a^*, \text{TPR}_{a}^{(b)} \approx \frac{|P_{tr}(y|x^{+}, a) > 0.5|}{N_{+, a}} \approx \text{TPR}_{a}^{(u)}
\end{equation}

In the case of high subgroup separability, Eq. (\ref{eq:tpr_sep}) and Eq. (\ref{eq:tpr_recovered}) demonstrate that TPR of the underdiagnosed group is directly affected by bias from the training set while other groups are mainly unaffected. Given this difference across groups, an appropriately selected group fairness metric may be able to identify the bias, in some cases even without access to an unbiased test set \cite{wachterBiasPreservationMachine2021}. On the other hand, when subgroup separability is low, this property does not hold. With non-separable groups (i.e. $P(a \vert x) \approx \frac{1}{|A|},  \forall a \in A$), a trained model will be unable to learn separate subgroup mappings, shown in Eq. (\ref{eq:non_separable}).

\begin{equation} \label{eq:non_separable}
    \hat{p}(y \vert x^+, a) \approx P_{tr}(y \vert x^+), \ \forall a \in A
\end{equation}

Equations (\ref{eq:underdiagnosis_total}) and (\ref{eq:non_separable}) imply that the performance of the trained model degrades for all groups. Returning to the example of TPR, Eq. (\ref{eq:tpr_non_separable}) represents performance degradation for all groups when separability is poor. In such situations, we expect performance degradation to be uniform across groups and thus not be detected by group fairness metrics. The severity of the degradation depends on both the proportion of corrupted labels in the underdiagnosed subgroup and the size of the underdiagnosed subgroup in the dataset.  

\begin{equation} \label{eq:tpr_non_separable}
    \text{TPR}_{a}^{(b)} \approx \frac{|P_{tr}(y|x^{+}, a) > 0.5|}{N_{+, a}} \le \frac{|P(y|x^{+}, a) > 0.5|}{N_{+, a}} \approx \text{TPR}_{a}^{(u)}, \forall a \in A
\end{equation}

We have derived the effect of underdiagnosis bias on classifier performance for the two extreme cases of high and low subgroup separability. In practice, subgroup separability for real-world datasets may vary continuously between these extremes. In Section \ref{sec:experiments}, we empirically investigate (i) how subgroup separability varies in the wild, (ii) how separability impacts performance for each group when underdiagnosis bias is added to the datasets, (iii) how models encode sensitive information in their representations.

\section{Experiments and Results} \label{sec:experiments}

We support our analysis with experiments on five datasets adapted from a subset of the MEDFAIR benchmark \cite{zongMEDFAIRBenchmarkingFairness2023}. We treat each dataset as a binary classification task (no-disease vs disease) with a binary subgroup label. For datasets with multiple sensitive attributes available, we investigate each individually, giving eleven dataset-attribute combinations. The datasets cover the modalities of skin dermatology \cite{tschandlHAM10000DatasetLarge2018,grohEvaluatingDeepNeural2021,grohTransparencyDermatologyImage2022}, fundus images \cite{kovalykPAPILADatasetFundus2022}, and chest X-ray \cite{irvinCheXpertLargeChest2019,johnsonMIMICCXRDeidentifiedPublicly2019}. We record summary statistics for the datasets used in the supplementary material (Table \ref{tab:ds_summary}), where we also provide access links (Table \ref{tab:ds_access}). Our architecture and hyperparameters are listed in Table \ref{tab:hyperparams}, adapted from the experiments in MEDFAIR.

\subsection*{Subgroup separability in the real world}

We begin by testing the premise of this article: subgroup separability varies across medical imaging settings. To measure subgroup separability, we train binary subgroup classifiers for each dataset-attribute combination. We use test-set area under receiver operating characteristic curve (AUC) as a proxy for separability, reporting results over ten random seeds in Table \ref{tab:subgroup_sep}. 

\begin{table}[ht] 
\caption{Separability of protected subgroups in real-world datasets, measured by test-set AUC of classifiers trained to predict the groups. Mean and standard deviation are reported over ten random seeds, with results sorted by ascending mean AUC. }\label{tab:subgroup_sep}
\ra{1.2}

\centering
\begin{tabular}{@{}lllllcrlccc@{}}

\toprule
\textbf{Dataset-Attribute} & \phantom{ab} & \textbf{Modality} & \phantom{ab} & \multicolumn{3}{c}{\textbf{Subgroups}} & \phantom{ab} & \multicolumn{3}{c}{\textbf{AUC}} \\ 
\cmidrule(r){5-7} \cmidrule(r){9-11}
 & & & & Group 0 & & Group 1 & & $\mu$ & \phantom{a} & $\sigma$\\\midrule
 PAPILA-Sex & & Fundus Image & & Male & & Female  & & 0.642 & & 0.057 \\
 HAM10000-Sex & & Skin Dermatology & & Male & & Female  & & 0.723 & & 0.015 \\
 HAM10000-Age & & Skin Dermatology & & $<60$ & & $\geq 60$  & & 0.803 & & 0.020 \\ 
 PAPILA-Age & & Fundus Image & & $<60$ & & $\geq 60$  & & 0.812 & & 0.046 \\ 
 Fitzpatrick17k-Skin & & Skin Dermatology & & I-III & & IV-VI & & 0.891 & & 0.010 \\ 
 CheXpert-Age & & Chest X-ray & & $<60$ & & $\geq 60$  & & 0.920 & & 0.003 \\ 
 MIMIC-Age & & Chest X-ray & & $<60$ & &  $\geq 60$  & & 0.930 & & 0.002 \\ 
 CheXpert-Race & & Chest X-ray & & White & &  Non-White  & & 0.936 & & 0.005 \\
 MIMIC-Race & & Chest X-ray & & White & &  Non-White  & & 0.951 & & 0.004 \\
 CheXpert-Sex & & Chest X-ray & & Male & &  Female  & & 0.980 & & 0.020 \\ 
 MIMIC-Sex & & Chest X-ray & & Male & & Female  & & 0.986 & & 0.008 \\
 \bottomrule
\end{tabular}
\end{table}

Some patterns are immediately noticeable from Table \ref{tab:subgroup_sep}. All attributes can be predicted from chest X-ray scans with $> 0.9$ AUC, implying that the modality encodes substantial information about patient identity. Age is consistently well predicted across all modalities, whereas separability of biological sex varies, with prediction of sex from fundus images being especially weak. Importantly, the wide range of AUC results $[0.642 \rightarrow 0.986]$ across the dataset-attribute combinations confirms our premise that subgroup separability varies substantially across medical imaging applications. 
\subsection*{Performance degradation under label bias}

\begin{figure}[ht]
    \centering
    \includegraphics[width=\textwidth]{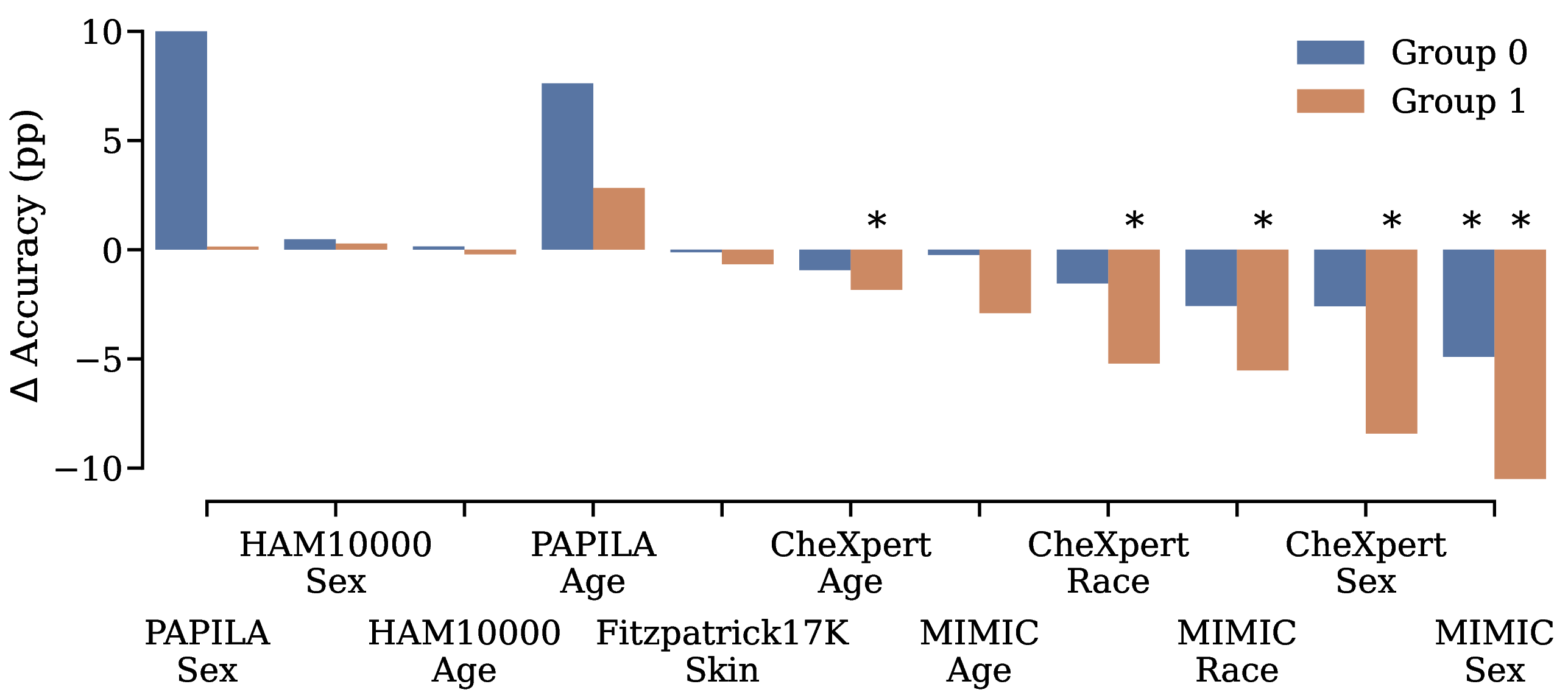}
    \caption{Percentage-point degradation in accuracy for disease classifiers trained on biased data, compared to training on clean data. Lower values indicate worse performance for the biased model when tested on a clean dataset. Results are reported over ten random seeds, and bars marked with $^*$ represent statistically significant results. Dataset-attribute combinations are sorted by ascending subgroup separability.}
    \label{fig:erm_degradation}
\end{figure}

We now test our theoretical finding: models are affected by underdiagnosis differently depending on subgroup separability. We inject underdiagnosis bias into each training dataset by randomly mislabelling $25 \%$ of positive individuals in Group 1 (see Table \ref{tab:subgroup_sep}) as negative. For each dataset-attribute combination, we train ten disease classification models with the biased training data and ten models with the original clean labels; we test all models on clean data. We assess how the test-time performance of the models trained on biased data degrades relative to models trained on clean data. We illustrate the mean percentage point accuracy degradation for each group in Fig. \ref{fig:erm_degradation} and use the Mann-Whitney U test (with the Holm-Bonferroni adjustment for multiple hypothesis testing) to determine if the performance degradation is statistically significant at $p_{\text{critical}}=0.05$. We include an ablation experiment over varying label noise intensity in Fig. \ref{fig:ln_ablation}. 

Our results in Fig. \ref{fig:erm_degradation} are consistent with our analysis in Section \ref{sec:analysis}. We report no statistically significant performance degradation for dataset-attribute combinations with low subgroup separability ($< 0.9 $ AUC). In these experiments, the proportion of mislabelled images is small relative to the total population; thus, the underdiagnosed subgroups mostly recover from label bias by sharing the correct mapping with the uncorrupted group. While we see surprising improvements in performance for PAPILA, note that this is the smallest dataset, and these improvements are not significant at $p_{\text{critical}}=0.05$. As subgroup separability increases, performance degrades more for the underdiagnosed group (Group 1), whilst performance for the uncorrupted group (Group 0) remains somewhat unharmed. We see a statistically significant performance drop for Group 0 in the MIMIC-Sex experiment -- we believe this is because the model learns separate group-wise mappings, shrinking the effective size of the dataset for Group 0.

\subsection*{Use of sensitive information in biased models}

\begin{figure}[ht]
    \centering
    \includegraphics[width=\textwidth]{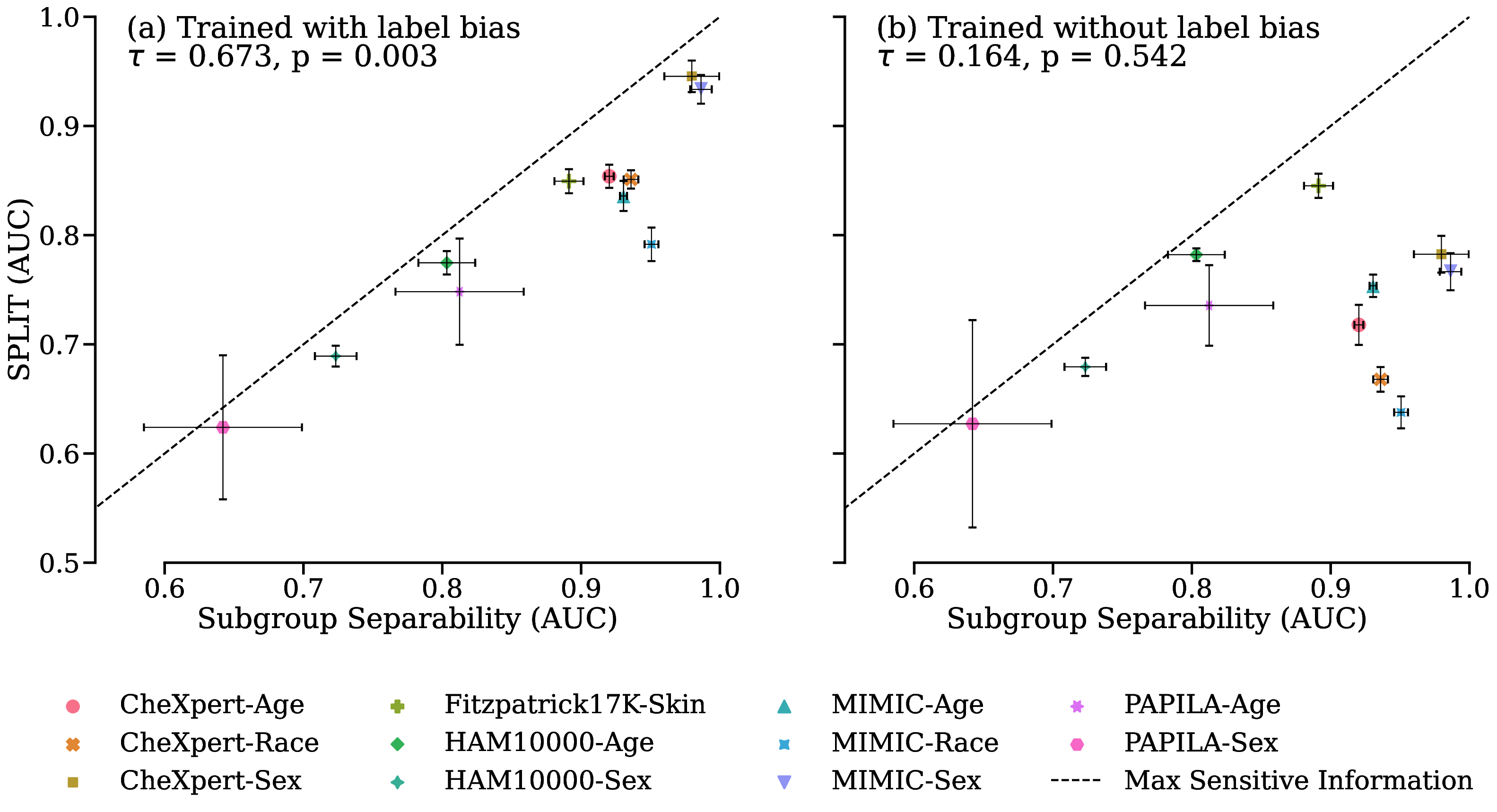}
    \caption{AUC of the SPLIT test for sensitive information encoded in learned representations, plotted against subgroup separability. Along the maximum sensitive information line, models trained for predicting the disease encode as much sensitive information in their representations as the images do themselves.}
    \label{fig:split_test}
\end{figure}

Finally, we investigate how biased models use sensitive information. We apply the post hoc Supervised Prediction Layer Information Test (SPLIT) \cite{glockerAlgorithmicEncodingProtected2023,gichoyaAIRecognitionPatient2022} to all models trained for the previous experiment, involving freezing the trained backbone and re-training the final layer to predict the sensitive attribute. We report test-set SPLIT AUC in Fig. \ref{fig:split_test}, plotting it against subgroup separability AUC from Table \ref{tab:subgroup_sep} and using Kendall's $\tau$ statistic to test for a monotonic association between the results ($p_{\text{critical}} = 0.05$). We find that models trained on biased data learn to encode sensitive information in their representations and see a statistically significant association between the amount of information available and the amount encoded in the representations. Models trained on unbiased data have no significant association, so do not appear to exploit sensitive information.


\section{Discussion}

We investigated how subgroup separability affects the performance of deep neural networks for disease classification. We discuss four takeaways from our study:  

\paragraph{Subgroup separability varies substantially in medical imaging.} In fairness literature, data is often assumed to contain sufficient information to identify individuals as subgroup members. But what if this information is only partially encoded in the data? By testing eleven dataset-attribute combinations across three medical modalities, we found that the ability of classifiers to predict sensitive attributes varies substantially. Our results are not exhaustive -- there are many modalities and sensitive attributes we did not consider -- however, by demonstrating a wide range of separability results across different attributes and modalities, we highlight a rarely considered property of medical image datasets.

\paragraph{Performance degradation is a function of subgroup separability.} We showed, theoretically and empirically, that the performance and fairness of models trained on biased data depends on subgroup separability. When separability is high, models learn to exploit the sensitive information and the bias is reflected by stark subgroup differences. When separability is low, models cannot exploit sensitive information, so they perform similarly for all groups. This indicates that group fairness metrics may be insufficient for detecting bias when separability is low. Our analysis centred on bias in classifiers trained with the standard approach of empirical risk minimisation -- future work may wish to investigate whether subgroup separability is a factor in the failure of bias mitigation methods and whether it remains relevant in further image analysis tasks (e.g. segmentation).

\paragraph{Sources of bias matter.} In our experiments, we injected underdiagnosis bias into the training set and treated the uncorrupted test set as an unbiased ground truth. However, this is not an endorsement of the quality of the data. At least some of the datasets may already contain an unknown amount of underdiagnosis bias (among other sources of bias) \cite{seyyed-kalantariUnderdiagnosisBiasArtificial2021,bernhardtPotentialSourcesDataset2022}. This pre-existing bias will likely have a smaller effect size than our artificial bias, so it should not play a significant role in our results. Still, the unmeasured bias may explain some variation in results across datasets. Future work should investigate how subgroup separability interacts with other sources of bias. We renew the call for future datasets to be released with patient metadata and multiple annotations to enable analysis of different sources and causes of bias.

\paragraph{Reproducibility and impact.} This work tackles social and technical problems in machine learning for medical imaging and is of interest to researchers and practitioners seeking to develop and deploy medical AI. Given the sensitive nature of this topic, and its potential impact, we have made considerable efforts to ensure full reproducibility of our results. All datasets used in this study are publicly available, with access links in Table \ref{tab:ds_access}. We provide a complete implementation of our preprocessing, experimentation, and analysis of results at \url{https://github.com/biomedia-mira/subgroup-separability}.

\subsection*{Acknowledgements}

 C.J. is supported by Microsoft Research and EPSRC through the Microsoft PhD Scholarship Programme. M.R. is funded through an Imperial College London President's PhD Scholarship. B.G. received support from the Royal Academy of Engineering as part of his Kheiron/RAEng Research Chair.

%

\bibliographystyle{splncs04}
\bibliography{references}

\begin{thebibliography}{10}
\providecommand{\url}[1]{\texttt{#1}}
\providecommand{\urlprefix}{URL }
\providecommand{\doi}[1]{https://doi.org/#1}

\bibitem{alviTurningBlindEye2018}
Alvi, M., Zisserman, A., Nellaaker, C.: Turning a {{Blind Eye}}: {{Explicit
  Removal}} of {{Biases}} and {{Variation}} from {{Deep Neural Network
  Embeddings}}. In: Proceedings of the {{European Conference}} on {{Computer
  Vision}} ({{ECCV}}) {{Workshops}} (Sep 2018)

\bibitem{bernhardtPotentialSourcesDataset2022}
Bernhardt, M., Jones, C., Glocker, B.: Potential sources of dataset bias
  complicate investigation of underdiagnosis by machine learning algorithms.
  Nature Medicine  \textbf{28}(6),  1157--1158 (Jun 2022).
  \doi{10.1038/s41591-022-01846-8}

\bibitem{brownDetectingPreventingShortcut}
Brown, A., Tomasev, N., Freyberg, J., Liu, Y., Karthikesalingam, A.: Detecting
  and {{Preventing Shortcut Learning}} for {{Fair Medical AI}} using {{Shortcut
  Testing}} ({{ShorT}})

\bibitem{castroCausalityMattersMedical2020}
Castro, D.C., Walker, I., Glocker, B.: Causality matters in medical imaging.
  Nature Communications  (2020). \doi{10.1038/s41467-020-17478-w}

\bibitem{degraveAIRadiographicCOVID192021}
DeGrave, A.J., Janizek, J.D., Lee, S.I.: {{AI}} for radiographic {{COVID-19}}
  detection selects shortcuts over signal. Nature Machine Intelligence
  \textbf{3}(7),  610--619 (Jul 2021). \doi{10.1038/s42256-021-00338-7}

\bibitem{geirhosShortcutLearningDeep2020}
Geirhos, R., Jacobsen, J.H., Michaelis, C., Zemel, R., Brendel, W., Bethge, M.,
  Wichmann, F.A.: Shortcut learning in deep neural networks. Nature Machine
  Intelligence  \textbf{2}(11),  665--673 (Nov 2020).
  \doi{10.1038/s42256-020-00257-z}

\bibitem{gichoyaAIRecognitionPatient2022}
Gichoya, J.W., Banerjee, I., Bhimireddy, A.R., Burns, J.L., Celi, L.A., Chen,
  L.C., Correa, R., Dullerud, N., Ghassemi, M., Huang, S.C., Kuo, P.C.,
  Lungren, M.P., Palmer, L.J., Price, B.J., Purkayastha, S., Pyrros, A.T.,
  {Oakden-Rayner}, L., Okechukwu, C., {Seyyed-Kalantari}, L., Trivedi, H.,
  Wang, R., Zaiman, Z., Zhang, H.: {{AI}} recognition of patient race in
  medical imaging: A modelling study. The Lancet Digital Health  \textbf{4}(6),
   e406--e414 (Jun 2022). \doi{10.1016/S2589-7500(22)00063-2}

\bibitem{glockerAlgorithmicEncodingProtected2023}
Glocker, B., Jones, C., Bernhardt, M., Winzeck, S.: Algorithmic encoding of
  protected characteristics in chest {{X-ray}} disease detection models.
  eBioMedicine  \textbf{89} (Mar 2023). \doi{10.1016/j.ebiom.2023.104467}

\bibitem{grohTransparencyDermatologyImage2022}
Groh, M., Harris, C., Daneshjou, R., Badri, O., Koochek, A.: Towards
  {{Transparency}} in {{Dermatology Image Datasets}} with {{Skin Tone
  Annotations}} by {{Experts}}, {{Crowds}}, and an {{Algorithm}}. Proceedings
  of the ACM on Human-Computer Interaction  \textbf{6}(CSCW2),  521:1--521:26
  (Nov 2022). \doi{10.1145/3555634}

\bibitem{grohEvaluatingDeepNeural2021}
Groh, M., Harris, C., Soenksen, L., Lau, F., Han, R., Kim, A., Koochek, A.,
  Badri, O.: Evaluating {{Deep Neural Networks Trained}} on {{Clinical Images}}
  in {{Dermatology With}} the {{Fitzpatrick}} 17k {{Dataset}}. In: Proceedings
  of the {{IEEE}}/{{CVF Conference}} on {{Computer Vision}} and {{Pattern
  Recognition}}. pp. 1820--1828 (2021)

\bibitem{irvinCheXpertLargeChest2019}
Irvin, J., Rajpurkar, P., Ko, M., Yu, Y., {Ciurea-Ilcus}, S., Chute, C.,
  Marklund, H., Haghgoo, B., Ball, R., Shpanskaya, K., Seekins, J., Mong, D.A.,
  Halabi, S.S., Sandberg, J.K., Jones, R., Larson, D.B., Langlotz, C.P., Patel,
  B.N., Lungren, M.P., Ng, A.Y.: {{CheXpert}}: {{A Large Chest Radiograph
  Dataset}} with {{Uncertainty Labels}} and {{Expert Comparison}}. Proceedings
  of the AAAI Conference on Artificial Intelligence  \textbf{33}(01),  590--597
  (Jul 2019). \doi{10.1609/aaai.v33i01.3301590}

\bibitem{jabbourDeepLearningApplied2020}
Jabbour, S., Fouhey, D., Kazerooni, E., Sjoding, M.W., Wiens, J.: Deep
  {{Learning Applied}} to {{Chest X-Rays}}: {{Exploiting}} and {{Preventing
  Shortcuts}}. In: Proceedings of the {{Machine Learning}} for {{Healthcare
  Conference}}. pp. 750--782. {PMLR} (Sep 2020)

\bibitem{johnsonMIMICCXRDeidentifiedPublicly2019}
Johnson, A.E.W., Pollard, T.J., Berkowitz, S.J., Greenbaum, N.R., Lungren,
  M.P., Deng, C.y., Mark, R.G., Horng, S.: {{MIMIC-CXR}}, a de-identified
  publicly available database of chest radiographs with free-text reports.
  Scientific Data  \textbf{6}(1), ~317 (Dec 2019).
  \doi{10.1038/s41597-019-0322-0}

\bibitem{kimLearningNotLearn2019}
Kim, B., Kim, H., Kim, K., Kim, S., Kim, J.: Learning not to learn:
  {{Training}} deep neural networks with biased data. In: Proceedings of the
  {{IEEE}}/{{CVF Conference}} on {{Computer Vision}} and {{Pattern
  Recognition}}. pp. 9012--9020 (2019)

\bibitem{kovalykPAPILADatasetFundus2022}
Kovalyk, O., {Morales-S{\'a}nchez}, J., {Verd{\'u}-Monedero}, R.,
  {Sell{\'e}s-Navarro}, I., {Palaz{\'o}n-Cabanes}, A., {Sancho-G{\'o}mez},
  J.L.: {{PAPILA}}: {{Dataset}} with fundus images and clinical data of both
  eyes of the same patient for glaucoma assessment. Scientific Data
  \textbf{9}(1), ~291 (Jun 2022). \doi{10.1038/s41597-022-01388-1}

\bibitem{mittelstadtUnfairnessFairMachine2023}
Mittelstadt, B., Wachter, S., Russell, C.: The {{Unfairness}} of {{Fair Machine
  Learning}}: {{Levelling}} down and strict egalitarianism by default (Jan
  2023)

\bibitem{nautaUncoveringCorrectingShortcut2021}
Nauta, M., Walsh, R., Dubowski, A., Seifert, C.: Uncovering and {{Correcting
  Shortcut Learning}} in {{Machine Learning Models}} for {{Skin Cancer
  Diagnosis}}. Diagnostics  \textbf{12}(1), ~40 (Dec 2021).
  \doi{10.3390/diagnostics12010040}

\bibitem{oakden-raynerHiddenStratificationCauses2020}
{Oakden-Rayner}, L., Dunnmon, J., Carneiro, G., R{\'e}, C.: Hidden
  {{Stratification Causes Clinically Meaningful Failures}} in {{Machine
  Learning}} for {{Medical Imaging}}. Proceedings of the ACM Conference on
  Health, Inference, and Learning  \textbf{2020},  151--159 (Apr 2020).
  \doi{10.1145/3368555.3384468}

\bibitem{Rajpurkar2017}
Rajpurkar, P., Irvin, J., Zhu, K., Yang, B., Mehta, H., Duan, T., Ding, D.,
  Bagul, A., Langlotz, C., Shpanskaya, K., Lungren, M.P., Ng, A.Y.:
  {{CheXNet}}: {{Radiologist-Level Pneumonia Detection}} on {{Chest X-Rays}}
  with {{Deep Learning}}  (Nov 2017)

\bibitem{seyyed-kalantariUnderdiagnosisBiasArtificial2021}
{Seyyed-Kalantari}, L., Zhang, H., McDermott, M.B., Chen, I.Y., Ghassemi, M.:
  Underdiagnosis bias of artificial intelligence algorithms applied to chest
  radiographs in under-served patient populations. Nature Medicine 2021 27:12
  \textbf{27}(12),  2176--2182 (Dec 2021). \doi{10.1038/s41591-021-01595-0}

\bibitem{tschandlHAM10000DatasetLarge2018}
Tschandl, P., Rosendahl, C., Kittler, H.: The {{HAM10000}} dataset, a large
  collection of multi-source dermatoscopic images of common pigmented skin
  lesions. Scientific Data  \textbf{5}(1),  180161 (Aug 2018).
  \doi{10.1038/sdata.2018.161}

\bibitem{vapnikOverviewStatisticalLearning1999}
Vapnik, V.: An overview of statistical learning theory. IEEE Transactions on
  Neural Networks  \textbf{10}(5),  988--999 (Sep 1999).
  \doi{10.1109/72.788640}

\bibitem{wachterBiasPreservationMachine2021}
Wachter, S., Mittelstadt, B., Russell, C.: Bias preservation in machine
  learning: The legality of fairness metrics under {{EU}} non-discrimination
  law. West Virginia Law Review  (2021)

\bibitem{wangFairnessVisualRecognition2020}
Wang, Z., Qinami, K., Karakozis, I.C., Genova, K., Nair, P., Hata, K.,
  Russakovsky, O.: Towards {{Fairness}} in {{Visual Recognition}}: {{Effective
  Strategies}} for {{Bias Mitigation}}. In: Proceedings of the {{IEEE}}/{{CVF
  Conference}} on {{Computer Vision}} and {{Pattern Recognition}} ({{CVPR}})
  (Jun 2020)

\bibitem{wilesFineGrainedAnalysisDistribution2022}
Wiles, O., Gowal, S., Stimberg, F., Rebuffi, S.A., Ktena, I., Dvijotham, K.D.,
  Cemgil, A.T.: A {{Fine-Grained Analysis}} on {{Distribution Shift}}. In:
  International {{Conference}} on {{Learning Representations}} (Jan 2022)

\bibitem{zietlowLevelingComputerVision2022}
Zietlow, D., Lohaus, M., Balakrishnan, G., Kleindessner, M., Locatello, F.,
  Sch{\"o}lkopf, B., Russell, C.: Leveling {{Down}} in {{Computer Vision}}:
  {{Pareto Inefficiencies}} in {{Fair Deep Classifiers}}. In: Proceedings of
  the {{IEEE}}/{{CVF Conference}} on {{Computer Vision}} and {{Pattern
  Recognition}}. pp. 10410--10421 (2022)

\bibitem{zongMEDFAIRBenchmarkingFairness2023}
Zong, Y., Yang, Y., Hospedales, T.: {{MEDFAIR}}: {{Benchmarking Fairness}} for
  {{Medical Imaging}}. In: {{International Conference}} on {{Learning
  Representations}} (Feb 2023)

\end{thebibliography}

\newpage
\appendix
\setcounter{table}{0}
\renewcommand{\thetable}{\Alph{section}\arabic{table}}
\setcounter{figure}{0}
\renewcommand{\thefigure}{\Alph{section}\arabic{figure}}

\section{Supplementary Material}

\begin{table}[ht] 
\caption{Summary statistics of datasets used in study. All datasets have a binary primary task, with disease labels (e.g. \textit{malignant}, \textit{pleural effusion}) binned into the positive class; non-disease labels (e.g. \textit{no-finding}, \textit{benign}) comprise the negative class. Percentages without brackets represent the proportion of images belonging to the subgroup. Percentages in brackets represent the prevalence of the positive class among the subgroup. N/A denotes no metadata available. Split percentages represent train/val/test.}\label{tab:ds_summary}
\ra{1.2}
\centering \scriptsize
\begin{tabular}{@{}lllllllllll@{}}
\toprule
 & \phantom{a} & \textbf{CheXpert} & \phantom{a} & \textbf{MIMIC} & \phantom{a} & \textbf{HAM10000} & \phantom{a} & \textbf{PAPILA} & \phantom{a} & \textbf{Fitzpatrick17k}  \\ 
\midrule
 \textbf{Images} & & 127118 & & 183207 & & 9958 & & 420 & & 16012 \\
 \textbf{Patients} & & 42884 & & 43209 & & N/A & & 210 & & N/A \\ 
 \textbf{Splits ($\%$)} & & 60/10/30 & & 60/10/30 & & 80/10/10 & & 70/10/20 & & 80/10/10 \\\midrule
 \textbf{Male} & & $58.8\%$ $ (91.6\%)$ & & $53.5\%$ $ (70.8\%)$ & & $54.2\%$ $ (16.8\%)$ & & $34.8\%$ $ (24.0\%)$ & & N/A \\
 \textbf{Female} & &  $41.2\%$ $ (91.2\%)$ & & $46.5\%$ $ (67.1\%)$ & & $45.8\%$ $ (11.6\%)$ & & $65.2\%$ $ (19.0\%)$ & & N/A \\ \midrule
 \textbf{Age $<60$} & & $39.2\%$ $ (87.1\%)$ & & $34.6\%$ $ (58.1\%)$ & & $71.9\%$ $ (9.55\%)$ & & $40.5\%$ $ (6.47\%)$ & & N/A \\
 \textbf{Age $\geq 60 $} & & $60.8\%$ $ (94.2\%)$ & & $65.4\%$ $ (74.9\%)$ & & $28.1\%$ $ (26.9\%)$ & & $59.5\%$ $ (30.4\%)$ & & N/A \\ \midrule
 \textbf{White} & & $77.9\%$ $ (91.7\%)$ & & $77.4\%$ $ (70.9\%)$ & & N/A & & N/A & & N/A \\
 \textbf{Non-White} & & $22.1\%$ $ (90.5\%)$ & & $22.6\%$ $ (62.7\%)$ & & N/A & & N/A & & N/A \\ \midrule
 \textbf{Skin I-III} & & N/A & & N/A & & N/A & & N/A & & $69.1\%$ $ (14.9\%)$ \\
 \textbf{Skin IV-VI} & & N/A & & N/A & & N/A & & N/A & & $30.9\%$ $ (10.3\%)$ \\
 
 \bottomrule
\end{tabular}

\end{table}

\begin{table}[ht] 
\caption{Access links for datasets used in the study.}\label{tab:ds_access}
\ra{1.2}
\centering \small
\begin{tabular}{@{}lll@{}}
\toprule
 \textbf{Dataset} & \phantom{ab} & \textbf{Access}  \\ \midrule
 \textbf{CheXpert} & & \url{https://stanfordmlgroup.github.io/competitions/chexpert/} \\
 \textbf{MIMIC} & & \url{https://physionet.org/content/mimic-cxr-jpg/2.0.0/} \\
 \textbf{HAM10000} & & \url{https://www.nature.com/articles/sdata2018161#Sec10} \\
 \textbf{PAPILA} & & \url{https://www.nature.com/articles/s41597-022-01388-1#Sec6} \\
 \textbf{Fitzpatrick17k} & & \url{https://github.com/mattgroh/fitzpatrick17k} \\
 
 \bottomrule
\end{tabular}

\end{table}

\begin{table}[ht] 
\caption{Hyperparameters and total compute used across all experiments.}\label{tab:hyperparams}
\ra{1.2}
\centering \small
\begin{tabular}{@{}lll@{}}
\toprule
 \textbf{Config} & \phantom{ab} & \textbf{Value} \\ \midrule
 \textbf{Architecture} & & ResNet18 \\
 \textbf{Optimiser} & & Adam \{lr: $2e-4$, $\beta_1$: $0.9$, $\beta_2$: $0.999$\} \\
 \textbf{LR Schedule} & & Constant \\
 \textbf{Max Epochs} & & $50$ \\
 \textbf{Early Stopping} & & \{Monitor: validation loss, Patience: $5$ epochs\}\\
 \textbf{Augmentation} & & RandomResizedCrop, RandomRotation($15^o$) \\
 \textbf{Batch Size} & & 256 (32 for PAPILA) \\ \midrule

\textbf{Total trained models} & & $ 495  = (330 \text{ main body} + 165 \text{ supplementary}) $ \\
 \textbf{Total Compute} & &  $\approx 100$ GPU Hours (NVIDIA RTX 3090 or equivalent) \\
 \textbf{Min GPU Memory} & & $\approx 8 \text{ GB}$ \\ 
 
 \bottomrule
\end{tabular}
\end{table}

\begin{figure}[ht]
    \centering
    \includegraphics[width=\textwidth]{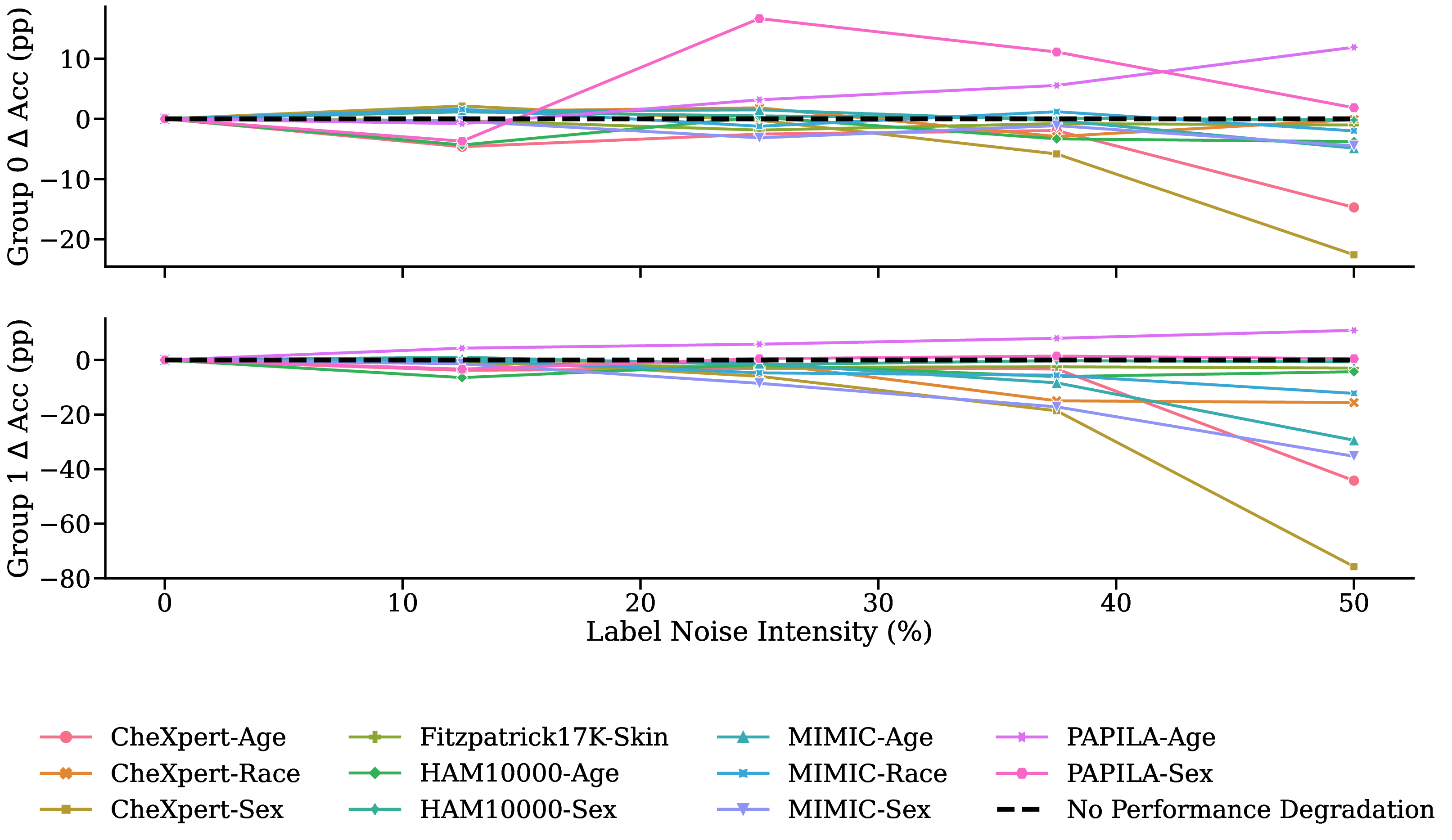}
    \caption{Percentage-point accuracy degradation with varying label noise intensity, where intensity is the percentage of mislabelled positive individuals in Group 1. Degradation is the difference in mean (over three random seeds) test-time accuracy for models trained with label noise compared to models trained without. The test dataset never contains label noise. Notice that the ordering of the datasets remains mostly consistent whilst label noise changes, showing that separability is predictive of performance degradation at all levels of label noise intensity. Group 1 performance degrades faster for datasets with high subgroup separability as label noise increases.}
    \label{fig:ln_ablation}
\end{figure}

\end{document}